\pdfoutput=1

\documentclass[11pt]{article}
\usepackage[]{acl}

\usepackage{times}
\usepackage{latexsym}

\usepackage{bbm}
\usepackage[russian, english]{babel}
\usepackage[T1]{fontenc}

\usepackage[utf8]{inputenc}

\usepackage[linesnumbered,ruled,vlined]{algorithm2e}
\SetKwInput{KwInput}{Input}   

\usepackage[utf8]{inputenc} 
\usepackage[T1]{fontenc}    
\usepackage{hyperref}       
\usepackage{url}            
\usepackage{amsfonts}       
\usepackage{nicefrac}       
\usepackage{microtype}      
\usepackage{xcolor}         

\usepackage{booktabs}
\usepackage{amsmath}
\usepackage{multirow}
\usepackage{graphicx}
\usepackage{enumitem}
\usepackage{subcaption}
\usepackage{caption}
\usepackage{rotating}
\usepackage{xcolor}
\newcommand{\ensuretext}[1]{#1}

\newcommand{\samarker}{\ensuretext{\textcolor{orange}{\ensuremath{^{\textsc{S}}_{\textsc{A}}}}}}

\newcommand{\reviewermarker}{\ensuretext{\textcolor{green}{\ensuremath{^{\textsc{R}}_{\textsc{W}}}}}}
\newcommand{\mycomment}[3]{}

\newcommand{\sa}[1]{\mycomment{\samarker}{#1}{orange}}

\newcommand{\rw}[1]{\mycomment{\reviewermarker}{#1}{green}}
\newcommand{\ignore}[1]{}

\usepackage[normalem]{ulem}
\usepackage{amssymb}
\usepackage{linguex}

\usepackage{siunitx} 
\usepackage{multirow, makecell}
\usepackage{pifont}
\usepackage{siunitx} 
\usepackage{multirow, makecell}
\usepackage{tabularx}
\usepackage[all]{nowidow}
\definecolor{darkgreen}{rgb}{0.0, 0.42, 0.24}
\definecolor{green}{RGB}{112, 173,71}
\definecolor{blue}{RGB}{68, 114,196}
\definecolor{orange}{RGB}{237, 125,49}
\definecolor{red}{RGB}{202, 54,49}
\definecolor{yellow}{RGB}{222,194, 142}

\definecolor{darkgreen}{RGB}{40, 54, 24}
\definecolor{offwhite}{RGB}{254, 250, 224}
\definecolor{darkorange}{RGB}{188, 108, 37}

\usepackage{xhfill}

\usepackage{xspace}
\usepackage{colortbl}
\usepackage{CJKutf8}

\newcommand{\mt}{\textsc{MT}\xspace}
\newcommand{\fsmt}{\textsc{FSMT}\xspace}
\newcommand{\nmt}{\textsc{NMT}\xspace}
\newcommand{\en}{\textsc{EN}\xspace}
\newcommand{\ita}{\textsc{IT}\xspace}
\newcommand{\ru}{\textsc{RU}\xspace}
\newcommand{\de}{\textsc{DE}\xspace}
\newcommand{\hi}{\textsc{HI}\xspace}
\newcommand{\ja}{\textsc{JA}\xspace}

\newcommand{\es}{\textsc{ES}\xspace}
\newcommand{\taskscorer}{\textsc{Acc.}\xspace}
\newcommand{\bert}{\textsc{BERT}\xspace}
\newcommand{\xlm}{\textsc{XLM-R}\xspace}
\newcommand{\mbart}{m\textsc{BART}-large\xspace}
\newcommand{\googlemt}{m\textsc{T5}-large\xspace}
\newcommand{\loss}{\mathcal{L}}

\newcommand{\bleu}{\textsc{BLEU}\xspace}
\newcommand{\ter}{\textsc{TER}\xspace}
\newcommand{\comet}{\textsc{COMET}\xspace}

\newcommand{\iwslt}{\textsc{IWSLT}\xspace}

\DeclareMathOperator*{\argmax}{arg\,max}

\newcommand{\internaltrain}{\textsc{Task train}\xspace}
\newcommand{\internaldev}{\textsc{Task dev}\xspace}

\usepackage{microtype}

\makeatletter
\newcommand{\printfnsymbol}[1]{%
  \textsuperscript{\@fnsymbol{#1}}%
}
\makeatother

\title{Controlling Translation Formality \\ Using Pre-trained Multilingual Language Models}

\author{Elijah Rippeth\thanks{\quad equal contribution.} \and \textbf{Sweta Agrawal\printfnsymbol{1}} \and  \textbf{Marine Carpuat} \\
         Department of Computer Science \\ University of Maryland \\
         \texttt{\{erip{,}sweagraw{,}marine\}@cs.umd.edu}}

\begin{document}
\maketitle
\begin{abstract}
This paper describes the University of Maryland's submission to the Special Task on Formality Control for Spoken Language Translation at \iwslt, which evaluates translation from English into 6 languages with diverse grammatical formality markers. We investigate to what extent this problem can be addressed with a \textit{single multilingual model}, simultaneously controlling its output for target language and formality. Results show that this strategy can approach the translation quality and formality control achieved by dedicated translation models. However, the nature of the underlying pre-trained language model and of the finetuning samples greatly impact results.
\end{abstract}

\section{Introduction}

While machine translation (\mt) research has primarily focused on preserving meaning across languages, linguists and lay-users alike have long known that 
pragmatic-preserving communication is an important aspect of the problem \cite{10.5555/913638}. To address one dimension of this, several works have attempted to control aspects of formality in \mt \cite{sennrich-etal-2016-controlling,feely-etal-2019-controlling,schioppa-etal-2021-controlling}. Indeed, this research area was formalized as formality-sensitive machine translation (\fsmt) by \citet{niu-etal-2017-study}, where the translation is not only a function of the source segment but also the desired target formality.
The lack of gold translation with alternate formality for supervised training and evaluation has lead researchers to rely on manual evaluation and synthetic supervision in past work \citep{niu2020controlling}. Additionally, these works broadly adapt to formal and informal registers 
as opposed to specifically controlling grammatical formality.

\begin{table}[t]
    \centering
        \renewcommand\tabularxcolumn[1]{m{#1}}
        \renewcommand\arraystretch{1.3}
    \begin{tabularx}{\columnwidth}{*{1}{>{\arraybackslash}X}}
  \textit{\textbf{Source:}} \textbf{Do you like}$_1$ Legos? \textbf{did you}$_2$ ever play with them as a child or even later? \\
  \textit{\textbf{German Informal:}} \textbf{Magst du}$_1$ Legos? \textbf{Hast du}$_2$ jemals als Kind mit ihnen gespielt oder sogar später? \\
  \textit{\textbf{German Formal:}} \textbf{Mögen Sie}$_1$ Legos? \textbf{Haben Sie}$_2$ jemals als Kind mit ihnen gespielt oder sogar später? \\
  
    \end{tabularx}  
    \caption{Contrastive formal and informal translations into German. Grammatical formality markers are bolded and aligned via indices.} \label{tab:example}
    \vspace{-0.7cm}
\end{table}

The Special Task on Formality Control on Spoken Language Translation provides a new benchmark by contributing high-quality training datasets for diverse languages \sa{Added citation for camera ready:} \cite{nadejde-etal-2022-coca-mt}. In this task, a source segment in English is paired with two references which are minimally contrastive in grammatical formality, one for each formality level (formal and informal; Table~\ref{tab:example}). Training and test samples are provided in the domains of ``telephony data'' and ``topical chat'' \cite{Gopalakrishnan2019} for four language pairs (English-\{German (\de), Spanish (\es), Hindi (\hi), Japanese(\ja)\}) and a test dataset for two additional ``zero-shot'' (ZS) language pairs (\en-\{Russian (\ru), Italian (\ita)\}). Markers of grammatical formality vary across these languages. Personal pronouns and verb agreement mark formality in many Indo-European languages (e.g., \de, \hi, \ita, \ru, \es), while in \ja, Korean (\textsc{KO}) and other languages, distinctions can be more extensive (e.g., using morphological markers) to express polite, respectful, and humble speech. 

In this work, we investigate how to control grammatical formality in \mt for many languages with minimal resources. Specifically, we ask whether a single multilingual model can be finetuned to translate in the appropriate formality for any of the task languages. We introduce additive vector interventions to encode style on top of \googlemt \cite{xue-etal-2021-mt5} and \mbart \cite{liu-etal-2020-multilingual-denoising}, and investigate the impact of finetuning on varying types of gold and synthetic samples to minimize reliance on manual annotation.


\section{Method}

Given an input sequence $x$, we design a \textit{single model} that produces an output $$y^{*} = \argmax p(y|x,l,f;\theta_{LM},\theta_{F})$$ for any language $l$ and formality level $f$ considered in this task. The bulk of its parameters $\theta_{LM}$ are initialized with a pre-trained multilingual language model. A small number of additional parameters $\theta_{F}$ enable formality control. All parameters are finetuned for formality-controlled translation.


\subsection{Multilingual Language Models}

We experiment with two underlying multilingual models: 1) \textbf{\googlemt}\footnote{24 layers with 1024 sized embeddings, 2816 FFN embedding dimension, and 16 heads for both encoder and decoder.} --- a multilingual variant of \textsc{T5} that is pre-trained on the Common Crawl-based dataset covering $101$ languages and 2) \textbf{\mbart}\footnote{12 layers with 1024 sized embeddings, 4096 FFN embedding dimension, and 16 heads for both encoder and decoder.} --- a Transformer encoder-decoder which supports multilingual machine translation for $50$ languages. While \mbart is pre-trained with parallel and monolingual supervision, \googlemt uses only monolingual dataset during the pre-training phase. 
Following standard practice, m\textsc{T5} controls the output language, $l$, via prompts (``Translate to German''), and m\textsc{BART} replaces the beginning of sequence token in the decoder with target language tags ($<$2xx$>$). 

\subsection{Additive Formality Control} \label{sec:covariate}

While large-scale pre-trained language models have shown tremendous success in multiple monolingual and multilingual controlled generation \cite{zhang2022survey} and style transfer tasks, their application to controlled cross-lingual text generation have been limited. Few-shot style-transfer approaches \cite{garcia-etal-2021-towards, riley-etal-2021-textsettr, krishna2022fewshot} hold the promise of minimal supervision but perform poorly on low-resource settings and their outputs lack diversity.

\begin{figure}[htb!]
\centering
  \includegraphics[width=\linewidth]{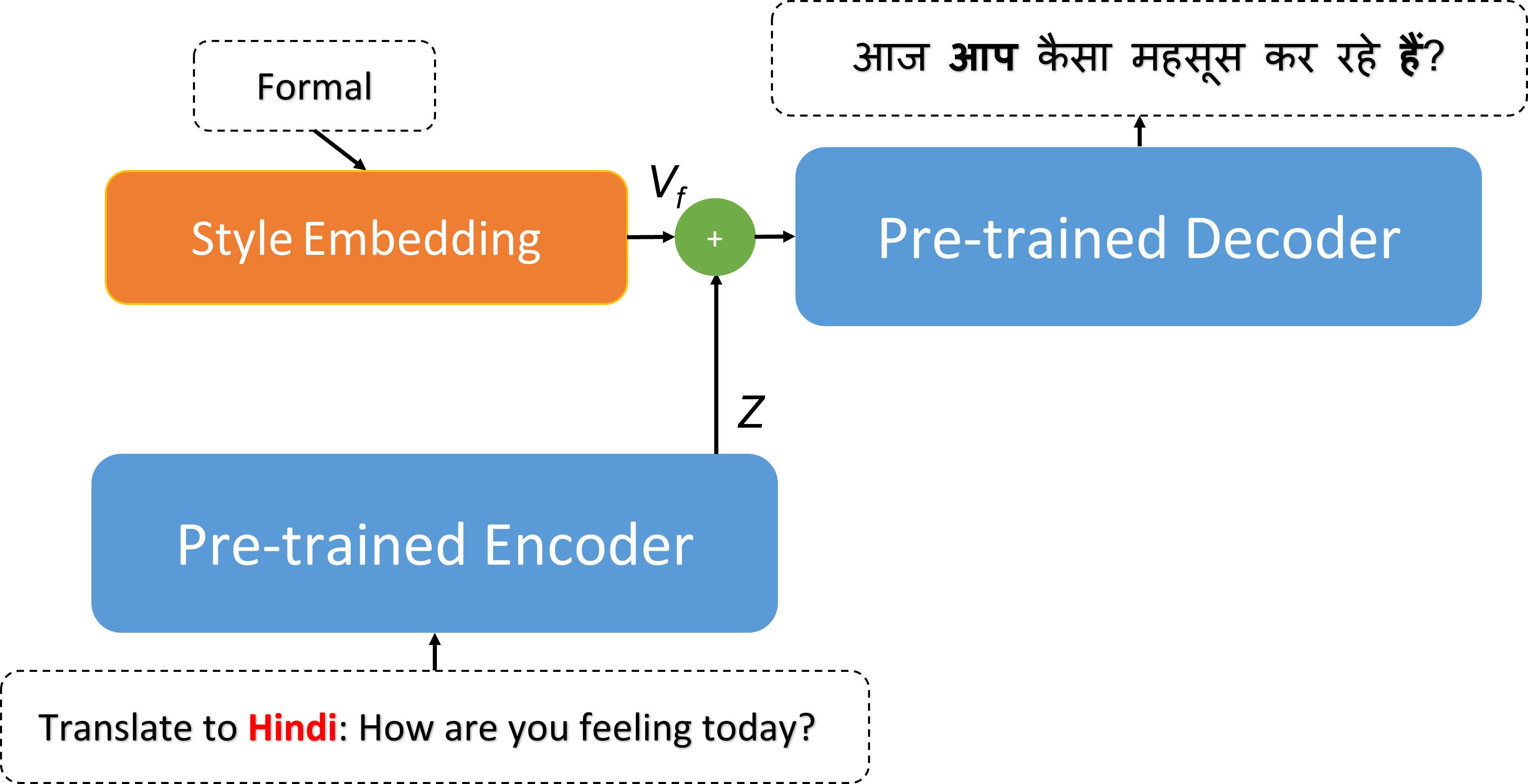}
    \caption{Controlling the output formality of a multilingual language model with additive interventions.}  \label{fig:main_covariate}
    \vspace{-0.5cm}
\end{figure}

\sa{Added for camera ready: } 
A popular way of introducing control when generating text with a particular style attribute is \textit{tagging}, where the desired control tags (e.g., $<$2formal$>$) are appended to the source or the target sequence. However, as discussed in \citet{schioppa-etal-2021-controlling}, this approach has several limitations, including but not limited to the necessity of including the control tokens in the vocabulary at the start of the training, which restricts the enhancement of pre-trained models with controllability. 

We introduce formality control by adapting the vector-valued interventions proposed by \citet{schioppa-etal-2021-controlling} for machine translation (\mt), as illustrated in Figure~\ref{fig:main_covariate}. 
%
Formally, given source text $x$, a formality level $f$, an encoder $E$ and decoder $D$, parameterized by $\theta_{LM}$, and a style embedding layer ($\text{Emb}$) parameterized by $\theta_F$ with the same output dimension as $E$, we have
\begin{align*}
    Z & = E(x){,} \quad V = \text{Emb}(f) \\
    y & = D(Z + V)
\end{align*}
Our formality levels can take values corresponding to formal, informal, and ``neutral'' translations, the last of which is used to generate ``generic'' translations in which there is no difference in the grammatical formality of the translation of the source if translated formally or informally. Unlike \citet{schioppa-etal-2021-controlling} who use a zero-vector as their neutral vector, we learn a separate vector.


\subsection{Finetuning}

Finetuning each multilingual model requires triplets of the form $(x, y, f)$ for each available target language, $l$, where $x$, $y$ and $f$ are the source text, the reference translation and the formality label corresponding to the reference translation respectively. The loss function is then given by:
\begin{equation}
\begin{split}\label{eq:loss}
\loss = \sum_{(x,y, l, f)} \log p(y|x,l,f;\theta_{LM},\theta_{F})
\end{split}
\end{equation}


\begin{table*}[t]
\centering
\begin{tabular}{lrrrrrrrr}
 \toprule
	Language &	\multicolumn{2}{c}{Size}  &	\multicolumn{3}{c}{Length}  & \multicolumn{3}{c}{Style}  \\
	& Train & Test &  Source & Formal & Informal & Avg. \ter  & \# Phrasal & \# Neutral \\
 \midrule
 \en-\de & 400& 600 & 22.78 & 24.68& 24.57& 0.126 & 1.89 & 23 \\
 \en-\es  & 400& 600&22.72 &22.64 &22.60 & 0.089& 1.56 &  49\\
 \en-\hi  & 400& 600 & 22.90& 25.92& 25.92& 0.068 & 1.57& 68\\
 \en-\ja  & 1000& 600& 24.61& 32.43&30.80 & 0.165 &2.47 & 20 \\
 \bottomrule
  \end{tabular}
 \caption{Shared Task Data Statistics: We use ``13a'' tokenization for all languages except Japanese for which we use ``ja-mecab'' implemented in the sacrebleu library.} \label{tab:data_stats}
\end{table*}

Given \textit{paired contrastive} training samples of the form $(X, Y_f, Y_{if})$, as provided by the shared task, the loss decomposes into balanced formal and informal components, but does not explicitly exploit the fact that $Y_i$ and $Y_f$ align to the same input:
\begin{equation}
\begin{split}\label{eq:loss}
\loss = \sum_{(x,y_f, l)} \log p(y_{f}|x,l,f;\theta_{LM},\theta_{F})  + \\ \sum_{(x,y_{if}, l)} \log p(y_{if}|x,l,if;\theta_{LM},\theta_{F})
\end{split}
\end{equation}

\subsection{Synthetic Supervision} \label{sec:synthetic}

Since paired contrastive samples are expensive to obtain, we explore the use of synthetic training samples to replace or complement them.
%
%
This can be done either by automatically annotating naturally occurring bitext for formality, which yields formal and informal samples, and additionally by rewriting the translation to alter its formality to obtain paired contrastive samples. The second approach was used by \citet{niu2020controlling} to control the register of \mt output. However, since this shared task targets grammatical formality and excludes other markers of formal vs. informal registers, we focus on the first approach, thus prioritizing control on the nature of the formality markers in the output over the tighter supervision provided by paired contrastive samples. 

Given a translation example $(x,y)$, we predict a silver-standard formality label ($f$) for the target $y$ using two distinct approaches:
\begin{itemize}[leftmargin=*]
    \item Rules (\es, \de, \ita, \ru): We label formality using heuristics based on keyword search, dependency parses, and morphological features. We use spaCy \cite{Honnibal_spaCy_Industrial-strength_Natural_2020} to (non-exhaustively) retrieve documents that imply a necessarily formal, necessarily informal, or ambiguously formal label. In the case of an ambiguously formal label, we treat it as unambiguously formal (for examples, see \ref{sec:glosses}). The complete set of rules for each of the  languages are included in the Appendix Table~\ref{tab:rules}. While simple to implement, these heuristics privilege precision over recall, and risk biasing the synthetic data to the few grammatical aspects they encode.
    \item Classifiers (\hi, \ja, \ita, \ru): We train a binary formal vs. informal classifier on the shared task data (\hi, \ja) and on the synthetic data (\ita, \ru). Unlike rules, they can also be transferred in a zero-shot fashion to new languages, and might be less biased toward precision when well-calibrated.
\end{itemize}

%

\section{Evaluation Settings}

\paragraph{Data} The shared task provides English source segments paired with two contrastive reference translations, one for each formality level (informal and formal) for four language pairs: \en-\{\de, \es, \ja, \textsc{HI}\} in the \textit{supervised} setting and two language pairs: \en-\{\ru, \textsc{IT}\} in the \textit{zero-shot} setting.\ The sizes and properties of the datasets for the supervised language pairs are listed in Table~\ref{tab:data_stats}. Formal texts tend to be longer and more diverse than informal texts for \ja compared to other language pairs. The percentage of neutral samples (same formal and informal outputs) vary from $2\%$ (in \ja) to $17\%$ (in \hi). In the \textit{zero-shot} setting, $600$ test samples are released for the two language pairs (\ru, \ita). 

During development, the last $50$ paired contrastive examples from each domain 
are set aside as a validation set for each of the supervised languages (\internaldev) and use the remaining samples for training (\internaltrain).

\begin{table*}[!htb]
    \centering
    \scalebox{0.80}{
    \begin{tabular}{lcrp{11cm}}
    \toprule
       \textbf{Model} &  \textbf{Target Language} &  \textbf{Size} & \textbf{Source} \\ \midrule
      \multirow{5}{*}{Synthetic Finetuned} & \ja  & {$15$K} & JParaCrawl \cite{morishita-etal-2020-jparacrawl} \\ 
      &  \hi & {$13$K} &  CCMatrix \cite{schwenk-etal-2021-ccmatrix} \\ 
      &  \ita, \ru & {$15$K} & Paracrawl v8 \cite{banon-etal-2020-paracrawl} \\ 
      &  \de & {$15$K} & CommonCrawl, Europarl v7 \cite{koehn-2005-europarl} \\ 
     &   \es & {$15$K} & CommonCrawl, Europarl v7, UN \cite{ziemski-etal-2016-united} \\ 
         \addlinespace[0.15cm]
   \multirow{3}{*}{Bilingual Baselines}   &   \de,\es,\ita,\ru & $20$M &  Paracrawl v9 \\
     &   \hi & $0.7$M  & CCMatrix \\
     &   \ja & $3.2$M & Wikimatrix \cite{schwenk-etal-2021-wikimatrix}, JESC \cite{pryzant-etal-2018-jesc}  \\
        \bottomrule
    \end{tabular}
    }
    \caption{Data sources from which unlabeled formality parallel examples are sampled for \en-X for training the \textit{Synthetic Finetuned} and the \textit{Bilingual} baselines. 
    }
\label{tab:data_sources} 
\end{table*}

\paragraph{Metrics} \label{metrics} We evaluate the translation quality of the detruecased detokenized outputs from each systems using \textbf{\bleu} \cite{papineni-etal-2002-bleu} and \textbf{\comet} \cite{rei-etal-2020-comet}. We use the \textsc{13a} tokenizer to report \textsc{sacreBLEU}\footnote{\url{https://pypi.org/project/sacrebleu/2.0.0/}} scores for all languages, except Japanese, for which we use the \textsc{ja-mecab}. 
We also report the official \textbf{formality accuracy} (\taskscorer). Given a set of hypotheses $H$, sets of corresponding phrase-annotated formal references $F$ and informal references $IF$, and a function $\phi$ yielding phrase-level contrastive terms from a reference, the task-specific evaluation metric is defined as follows:
\begin{align*} \label{eq:formality}
    match_f & = \sum_j \mathbbm{1}\left[\phi(F_j) \in H_j \wedge \phi(IF_j) \notin H_j\right] \\
    match_i & = \sum_j \mathbbm{1}\left[\phi(F_j) \notin H_j \wedge \phi(IF_j) \in H_j\right] \\
    acc_j & = \frac{match_j}{match_f + match_{i}}, \quad j \in \{f, i\}
\end{align*}
We note that the task accuracy is a function of the number of \textit{matches} in the hypotheses, not the number of \textit{expected} phrases, i.e. ${match_f + match_{if}} \leq \|H\|$ and discuss the implications in the Appendix (Section~\ref{sec:eval}). 

\section{Experimental Conditions}

We compare multilingual models, where a single model is used to generate formal and informal translations for all languages with bilingual models trained for each language pair, as detailed below.

\subsection{Multilingual Models}


\paragraph{Data} We consider three finetuning settings:

\begin{itemize}[leftmargin=*]
\item \textbf{Gold finetuned}: the model is finetuned only on \textit{paired contrastive} shared task data (400 to 1000 samples per language pair).
\item \textbf{Synthetic finetuned}: the model is finetuned on \textit{synthetic silver-labelled triplets} (up to 7500 samples per formality level and language as described below). 
\item \textbf{Two-pass finetuned}: the \textit{Synthetic finetuned} model is further finetuned on a mixture of gold data and 1000 examples re-sampled from the synthetic training set for unseen languages, which we use to avoid catastrophic forgetting from the silver finetuning stage.
\end{itemize}

Synthetic samples are drawn from multiple data sources (\ref{tab:data_sources}), sampling at most $7500$ examples for each language and formality level. \sa{Footnote Added for camera ready: }  \footnote{We do not experiment with varying the sizes of the synthetic dataset due to the time constraints and leave it to the future work.} The formality labels are predicted as described in \ref{sec:synthetic}. Rule-based predictors directly give a label. With classifiers, we assign the formal label if $P(\text{formal}|y)\geq0.85$ and informal if $P(\text{formal}|y)\leq0.15$.

We additionally compare with the translations generated from the base \mbart model with no finetuning, referred to as the ``\textit{formality agnostic \mbart}''.

\paragraph{Training settings} We finetune \googlemt and \mbart with a batch size of $2$ and $8$ respectively for $10$ and $3$ epochs respectively. We mask the formality labels used to generate vector-valued interventions with a probability of $0.2$. The \googlemt model --- ``\textit{synthetic finetuned \googlemt}'' --- is trained for an additional $5$ epochs, with a batch size of $2$ on a mixture of task data for seen languages and a subset of the sampled synthetic data for unseen languages. Again, we mask the formality tag with probability $0.2$ except in the case of unseen languages where the formality tag is masked with probability $1.0$, resulting in the ``\textit{two-pass finetuned \googlemt}'' model.

\begin{table*}[h]
\centering
 \setlength\tabcolsep{2pt}
\scalebox{0.9}{
\begin{tabular}{lccrrrrrrrrr}
 \toprule
  \multirow{2}{*}{\textbf{\textsc{Samples}}} & \multirow{2}{*}{\textbf{\textsc{To}}}  && \multicolumn{2}{c}{\textbf{\en-\de}}  & \multicolumn{2}{c}{\textbf{\en-\hi}} & \multicolumn{2}{c}{\textbf{\en-\ja}}  &  \multicolumn{2}{c}{\textbf{\en-\es}}  \\
  & &&  \bleu & \taskscorer & \bleu  & \taskscorer & \bleu & \taskscorer & \bleu & \taskscorer \\
     \midrule
     \addlinespace[0.2cm]
Paired Contrastive & \textsc{F}&& 35.0 & \textbf{100}  & 28.7 & 98.7 & 33.1 & 95.3 & 32.6 & \textbf{100} \\
 
Unpaired Triplets & \textsc{F} && \textbf{35.5} & \textbf{100} & \textbf{31.6} & \textbf{100} & \textbf{39.6} & \textbf{100} & \textbf{35.5} & \textbf{100}\\

\addlinespace[0.2cm]
 Paired Contrastive & \textsc{IF} && 32.7 & 98.5 & 26.4 & 98.3 & 32.3 & \textbf{100} & 33.8 & \textbf{100} \\
 
Unpaired Triplets & \textsc{IF} && \textbf{35.9} & \textbf{98.6} & \textbf{30.9} & \textbf{98.4} & \textbf{40.3} & \textbf{100} & \textbf{39.6} & 97.9\\
\bottomrule
 \end{tabular}
 }
\caption{Results on  the \internaldev split when training \textit{Additive \googlemt} with and without contrastive examples: Sample diversity from Unpaired triplets improve \bleu and Accuracy over paired contrastive samples.}
\label{tab:paired_unpaired} 
\end{table*}


\begin{table}[h]
\centering
 \setlength\tabcolsep{2pt}
\scalebox{0.9}{
\begin{tabular}{lrrrr}
 \toprule
 \textbf{\textsc{Data}} & \textbf{\en-\de}  & \textbf{\en-\hi} & \textbf{\en-\ja}  &  \textbf{\en-\es}  \\
     \midrule
   Paired Contrastive &  0.397 & 0.371 & 0.421 & 0.505\\
 Unpaired Triplets & 0.459 & 0.415 & 0.460 & 0.580\\
\bottomrule
 \end{tabular}
 }
\caption{Results on the \internaldev split: \ter between generated formal and informal sentences.}
\vspace{-0.5cm}
\label{tab:paired_unpaired_TER} 
\end{table}


\paragraph{Formality Classifiers}\label{subsec:formality} Following \citet{briakou-etal-2021-evaluating}, we finetune \xlm on binary classification between formal and informal classes, using the shared task datasets for each of the supervised language pairs (\de, \es, \ja, \hi) and synthetic datasets for zero-shot language pairs (\ru, \ita). \sa{Added for camera ready: } We treat the ``neutral'' samples as both ``formal'' and ``informal'' when training the classifiers. We use the Adam optimizer, a batch size of $32$, and a learning rate of $5\times10^{-3}$ to finetune for $3$ epochs. We report the accuracy of the learned classifiers trained on the \internaltrain dataset in Appendix Table~\ref{tab:accuracy}. 

\subsection{Bilingual Models}

We consider two types of bilingual models:
\begin{enumerate}[leftmargin=*]
    \item \textbf{Formality Agnostic:} These models were released by the shared task organizers. Each model is bilingual and trained on a sample of $20$ million lines from the Paracrawl Corpus (V9) using the Sockeye \nmt toolkit. Models use big transformers with $20$ encoder layers, $2$ decoder layers, SSRU's in place of decoder self-attention, and large batch training. 
    \item \textbf{Formality Specific (Gold):} We finetune the models in [1] to generate a formal model and an informal model for each language pair (except the zero-shot language pairs).
\end{enumerate}

The effective capacity of the bilingual, formality specific models is $3.14$B parameters.
Each model has $314$M parameters, resulting in $(314\times2\times4)=2.5$B parameters for the four supervised languages (\de, \es, \hi, \ja) and two pre-trained models $(314\times2)=628$M parameters for the unseen languages (\ru, \ita).
This is significantly larger than the capacities of our single multilingual models (Additive \googlemt: $1.25$B, Additive \mbart: $610$M). 

%


\begin{table*}[t]
\centering
 \setlength\tabcolsep{2pt}
\scalebox{0.80}{
\begin{tabular}{lrrrrrrrrrrrrr}
 \toprule
  \multirow{2}{*}{\textbf{\textsc{Model}}} & \multicolumn{3}{c}{\textbf{\en-\de}}  & \multicolumn{3}{c}{\textbf{\en-\es}} & \multicolumn{3}{c}{\textbf{\en-\ja}}  &  \multicolumn{3}{c}{\textbf{\en-\hi}}  \\
  & \bleu & \comet & \taskscorer & \bleu & \comet & \taskscorer & \bleu & \comet & \taskscorer & \bleu & \comet & \taskscorer \\
     \midrule
 \rowcolor{gray!10}
  \textbf{\textit{Bilingual}}  \\
  Formality Agnostic & 33.2 & 0.432 & 33.8 & 41.3 & 0.675 & 24.5 & 13.0 &  -0.093 &  25.6 & 27.8 & 0.464&  96.5 \\
  Formality Specific (Gold)  & 49.1 & 0.539 & 100.0 & 56.0 & 0.790 & 100.0 & 26.0 & 0.242 & 89.1 & 37.5 &0.694& 100.0 \\
      \midrule
        \addlinespace[0.1cm]
  \textbf{\textit{Multilingual Model}}  \\
  \addlinespace[0.1cm]
 \rowcolor{gray!10}
\textit{mBART-large}  \\
 Formality Agnostic & 33.3 & 0.295 & 68.9 & 27.0 &  0.120& 56.5 & 18.3 & -0.016 & 71.9 & 20.7 &0.340& 88.4 \\
Gold Finetuned & 42.8 & 0.462 & 95.9 & 41.1 & 0.548& 97.7 & 24.7 & 0.326& 89.4 & 29.6 & 0.678& 95.6 \\

  \addlinespace[0.1cm]
 \rowcolor{gray!10}
 \textit{mT5-large}  \\
Gold Finetuned &  53.3 &0.260& \textbf{100.0} & 53.5 & 0.427& \textbf{100.0} & 49.8 & 0.645& 98.1 & 43.5 & 0.359& \textbf{100.0}\\
 Synthetic Finetuned & 64.5 & 0.557& \textbf{100.0} & 50.7 & 0.345 & \textbf{100.0} & 58.5 & 0.837 & 97.7 &61.2 & 0.844 & \textbf{100.0}\\
Two-pass Finetuned & \textbf{86.8} &\textbf{0.824}& \textbf{100.0} & \textbf{88.2} &\textbf{1.070}& \textbf{100.0} & \textbf{68.3} & \textbf{0.980}& \textbf{100.0} & \textbf{70.4} & \textbf{0.975}& \textbf{100.0}\\

\bottomrule
 \end{tabular}
 }
\caption{Results on the \internaldev split in the \textit{\textbf{formal}} \textit{supervised} setting. \taskscorer: \textit{formal} accuracy.}
\label{tab:unofficial_formal_supervised} 
\end{table*}

\begin{table*}[t]
\centering
 \setlength\tabcolsep{2pt}
\scalebox{0.80}{
\begin{tabular}{lrrrrrrrrrrrrr}
 \toprule
  \multirow{2}{*}{\textbf{\textsc{Model}}} & \multicolumn{3}{c}{\textbf{\en-\de}}  & \multicolumn{3}{c}{\textbf{\en-\es}} & \multicolumn{3}{c}{\textbf{\en-\ja}}  &  \multicolumn{3}{c}{\textbf{\en-\hi}}  \\
  & \bleu & \comet & \taskscorer & \bleu & \comet & \taskscorer & \bleu & \comet & \taskscorer & \bleu & \comet & \taskscorer \\
     \midrule
  \textbf{\textit{Bilingual}}  \\
 Formality Agnostic & 37.2 & 0.470 & 66.2 & 45.8 & 0.691& 75.5 & 13.5 & -0.096 & 74.4 & 23.7 &0.436& 3.5 \\
   Formality Specific (Gold)  & 48.4 & 0.557& 98.5 & 55.1 &0.813& 95.7 & 22.6 &0.182& 97.8 & 36.3 &0.675& 91.5 \\
     \midrule
        \addlinespace[0.1cm]
  \textbf{\textit{Multilingual Model}}  \\
  \addlinespace[0.1cm]
 \rowcolor{gray!10}
\textit{mBART-large}  \\
   
  Formality Agnostic & 29.3 & 0.262& 31.1 & 26.3 &0.101& 43.5 & 16.2 & -0.036 & 28.1 & 18.7 &0.330& 11.6 \\
 Gold Finetuned & 39.6 & 0.456& 76.5 & 40.4 & 0.582& 95.3 & 21.6 & 0.289& 72.7 & 27.7 & 0.631& 82.8 \\

  \addlinespace[0.1cm]
 \rowcolor{gray!10}
 \textit{mT5-large}  \\
 
Gold Finetuned & 52.8 &0.232& \textbf{100.0} & 53.8 &0.513& \textbf{100.0} & 47.3 &0.617& \textbf{100.0} & 41.7 & 0.144& \textbf{100.0}\\
Synthetic Finetuned & 66.0 & 0.563& \textbf{100.0} & 57.6 & 0.530 & \textbf{100.0} &  59.0 & 0.813& 98.5 & 57.7 & 0.761 & \textbf{100.0}\\
 Two-pass Finetuned & \textbf{86.6} &\textbf{0.843}& \textbf{100.0} & \textbf{87.7} &\textbf{1.081}& \textbf{100.0} & \textbf{69.5} & \textbf{0.976}& \textbf{100.0} & \textbf{70.1} & \textbf{0.957}& \textbf{100.0}\\

\bottomrule
 \end{tabular}
 }
\caption{Results on  the \internaldev split in the \textit{\textbf{informal}} \textit{supervised} setting. \taskscorer: \textit{informal} accuracy.}
\label{tab:unofficial_informal_supervised}
\end{table*}

\section{System Development Results}


During system development, we explore the impact of different types of training samples and fine-tuning strategies on translation quality and formality accuracy on \internaldev.

\paragraph{Contrastive Samples} We estimate the benefits of fine-tuning on informal vs. formal translations of the same inputs for this task. We train two variants of the {\ttfamily gold finetuned \googlemt} model using $50$\% of the paired contrastive samples and $100\%$ of the unpaired triplets (i.e., selecting one formality level per unique source sentence) from the \internaltrain samples (Table~\ref{tab:paired_unpaired}). Results show that sample diversity resulting from unpaired triplets leads to better translation quality as measured by \bleu (Average Gain: Formal $+3.2$. Informal $+5.38$),  without compromising on the formality accuracy. Training with paired samples result in lower \ter between formal and informal output compared to unpaired triplets (Table~\ref{tab:paired_unpaired_TER}), suggesting that the outputs generated by the model trained on paired samples are more contrastive. This further motivates our {\ttfamily two-pass finetuned} model which uses gold contrastive samples on the final stage of finetuning to bias the model towards generating contrastive \mt outputs. 

\paragraph{}  

 While \internaldev is too small to make definitive claims, we report our system development results in  Tables~\ref{tab:unofficial_formal_supervised} and \ref{tab:unofficial_informal_supervised}. We observe that finetuning on gold contrastive examples  ({\ttfamily gold-finetuned}) improves the translation quality and accuracy of the translation models ({\ttfamily formality-agnostic}), highlighting the importance of limited but high-quality in-domain supervision on the resulting models. Further, each of the {\ttfamily \googlemt} models improves in translation quality with additional data and training. While the results are dramatic due to size of both \internaltrain and \internaldev, the trends validate the approach to augment both \mbart and the \googlemt with additive interventions to control formality.

\begin{table*}[h]
\centering
 \setlength\tabcolsep{1pt}
\scalebox{0.83}{
\begin{tabular}{llrcrrcrrcrrcr}
 \toprule
 &  \multirow{2}{*}{\textbf{\textsc{}}} & \multicolumn{3}{c}{\textbf{\en-\de}}  & \multicolumn{3}{c}{\textbf{\en-\es}} & \multicolumn{3}{c}{\textbf{\en-\ja}}  &  \multicolumn{3}{c}{\textbf{\en-\hi}}\\
 &  & \bleu & \comet & \taskscorer & \bleu & \comet & \taskscorer & \bleu & \comet & \taskscorer & \bleu & \comet & \taskscorer  \\
     \midrule
 &  \textbf{\textit{Bilingual Models}}  \\
&  Formality Agnostic & 33.0 & 0.472 & 53.6 & 37.5 & 0.646 & 37.9 & 14.9 & -0.102 & 23.3 & \textbf{26.5} & \textbf{0.519} & 98.8 \\
&  Formal Gold Finetuned &  \textbf{45.9} & \textbf{0.557} & \textbf{100.0} & \textbf{48.6} & \textbf{0.734} & \textbf{98.4} & \textbf{26.0} & \textbf{0.290} & \textbf{87.1} & 23.0 & 0.303 & \textbf{98.9} \\
     \midrule
        \addlinespace[0.1cm]
 &  \textbf{\textit{Multilingual Models}}  \\
  \addlinespace[0.1cm]
 \rowcolor{gray!10}
& \textit{mBART-large}  \\
 & Formality Agnostic & 35.1 & 0.344 & 83.6 & 26.9 & 0.210 & 67.8 & 18.3 &  0.051 & \textbf{93.4} & 20.1 & 0.383 & 93.5  \\ 

\textbf{\texttt{[4]}} & Gold Finetuned & \textbf{38.6} & \textbf{0.484} & 93.6 & 38.3 & \textbf{0.549} & 96.7 & \textbf{26.1} & \textbf{0.397} & 78.2 & 29.7 & \textbf{0.691} & 98.5  \\ 

  \addlinespace[0.1cm]
 \rowcolor{gray!10}
&  \textit{mT5-large}  \\
\textbf{\texttt{[3]}} & Gold Finetuned & 7.9 & -1.472 & \textbf{100.0} & 5.2 & -1.340 & 97.0 & 8.9 & -0.792 & 88.5 & 3.9 & -1.152 & \textbf{99.6} \\
\textbf{\texttt{[2]}} & Synthetic Finetuned  & 22.1 & 0.076 & 92.4 & 28.1 & 0.274 & 86.5 & 16.3 & -0.086 & 84.5 & 22.6 & 0.305 & 99.5\\
\textbf{\texttt{[1]}} & Two-pass Finetuned & 37.0 & 0.302 & 99.4 & \textbf{38.6} & 0.509 & \textbf{99.5} & 24.7 & 0.273 & 86.3 & \textbf{29.9} & 0.471 & 99.4\\
\bottomrule
 \end{tabular}
 }
\caption{Results on the official test split in the \textit{\textbf{formal}} \textit{supervised} setting. Best scores from multilingual and bilingual systems are \textbf{bolded}. Our official submissions to the shared task are numbered \textbf{\texttt{[1-4]}}.}
\label{tab:official_formal} 
\end{table*}

\begin{table*}[h]
\centering
 \setlength\tabcolsep{1pt}
\scalebox{0.83}{
\begin{tabular}{llrcrrcrrcrrcr}
 \toprule
&  \multirow{2}{*}{\textbf{\textsc{Model}}} & \multicolumn{3}{c}{\textbf{\en-\de}}  & \multicolumn{3}{c}{\textbf{\en-\es}} & \multicolumn{3}{c}{\textbf{\en-\ja}}  &  \multicolumn{3}{c}{\textbf{\en-\hi}} \\
 & & \bleu & \comet & \taskscorer & \bleu & \comet & \taskscorer & \bleu & \comet & \taskscorer & \bleu & \comet & \taskscorer \\
     \midrule
  &   \textbf{\textit{Bilingual Models}}  \\
& Formality Agnostic &  32.3 & 0.476 &  46.4 & 40.4 & 0.672 & 62.1  & 15.5 & -0.094 & 76.7 & 20.8 & 0.493 & 1.2  \\
 &  Formality Specific (Gold)  & \textbf{43.5} & \textbf{0.559} & \textbf{90.0} & \textbf{48.2} & \textbf{0.762} & \textbf{92.9}  & \textbf{23.5} & \textbf{0.272} & \textbf{98.7} & \textbf{31.2} & \textbf{0.724} & \textbf{92.1} \\
 \midrule
   \addlinespace[0.1cm]
  & \textbf{\textit{Multilingual Models}}  \\
  \addlinespace[0.1cm]
 \rowcolor{gray!10}
&  \textit{\mbart}  \\
 & Formality Agnostic & 28.4 & 0.299 & 16.4 & 25.3 & 0.205 & 32.2 & 16.2 & 0.032 & 6.6 & 16.7 & 0.370 & 6.5  \\ 

 \textbf{\texttt{[4]}} & Gold Finetuned & \textbf{36.1} & \textbf{0.472} & 77.4 & \textbf{38.3} & \textbf{0.549} & 82.7 & \textbf{22.8} & \textbf{0.346} & 88.0 & 27.6 & \textbf{0.670} & 64.7  \\
 
   \addlinespace[0.1cm]
 \rowcolor{gray!10}
&  \textit{\googlemt}  \\
 \textbf{\texttt{[3]}} & Gold Finetuned & 7.3 & -1.424 & 96.0 & 5.9 & -1.295 & \textbf{96.1} & 7.2 & -0.795 & \textbf{98.9} & 2.7 & -1.205 & 96.5  \\
\textbf{\texttt{[2]}} & Synthetic Finetuned  & 21.7 & 0.046 & 91.4 & 28.2 & 0.337 & 91.6 & 13.6 & -0.135 & 83.3 & 17.8 & 0.277 & 8.3 \\
\textbf{\texttt{[1]}} & Two-pass Finetuned & 35.9 & 0.301 & \textbf{96.5} & 38.0 & 0.539 & 93.2 & 22.3 & 0.252 & 97.5 & \textbf{29.2} & 0.439 & \textbf{98.7} \\
\bottomrule
 \end{tabular}
 }
\caption{Results on  the official test split in the \textit{\textbf{informal}} \textit{supervised} setting. Best scores from multilingual and bilingual systems are \textbf{bolded}. Our official submissions to the shared task are numbered \textbf{\texttt{[1-4]}}.}
\label{tab:official_informal} 
\end{table*}

\begin{table*}[h]
\centering
 \setlength\tabcolsep{1pt}
\scalebox{0.83}{
\begin{tabular}{llrcrrcrrcrrcr}
 \toprule
&  \multirow{2}{*}{\textbf{\textsc{Model}}} & \multicolumn{3}{c}{\textbf{\en-\de}}  & \multicolumn{3}{c}{\textbf{\en-\es}} & \multicolumn{3}{c}{\textbf{\en-\ja}}  &  \multicolumn{3}{c}{\textbf{\en-\hi}} \\
 & & \bleu & \comet & \taskscorer & \bleu & \comet & \taskscorer & \bleu & \comet & \taskscorer & \bleu & \comet & \taskscorer \\
     \midrule
  &   \textbf{\textit{Bilingual Models}}  \\
& Formality Agnostic &  32.7 & 0.474 & 50.0 & 39.0 & 0.659 & 50.0 & 15.2 & -0.100 & 50.0 & 23.7 & 0.506 & 50.0  \\
 &  Formality Specific (Gold)  & \textbf{ 44.7} & \textbf{0.558} & \textbf{95.0} & \textbf{48.4} & \textbf{0.748} & \textbf{95.7} & \textbf{24.8} & \textbf{0.281} & \textbf{92.9} & \textbf{27.1} & \textbf{0.513} & \textbf{95.5} \\
 \midrule
   \addlinespace[0.1cm]
  & \textbf{\textit{Multilingual Models}}  \\
  \addlinespace[0.1cm]
 \rowcolor{gray!10}
&  \textit{\mbart}  \\
 & Formality Agnostic & 31.8 & 0.322 & 50.0 & 26.1 & 0.207 & 50.0 & 17.3 & 0.041 & 50.0 & 18.4 & 0.377 & 50.0  \\ 

 \textbf{\texttt{[4]}} & Gold Finetuned & \textbf{37.4} & \textbf{0.478} & 85.5 & \textbf{38.3} & \textbf{0.549} & 89.7 & \textbf{24.5} & \textbf{0.371} & 83.1 & 28.7 & \textbf{0.680} & 81.6 \\
 
   \addlinespace[0.1cm]
 \rowcolor{gray!10}
&  \textit{\googlemt}  \\
 \textbf{\texttt{[3]}} & Gold Finetuned & 7.6 & -1.448 & \textbf{98.0} & 5.6 & -1.317 & \textbf{96.6} & 8.1 & -0.794 & \textbf{93.7} & 3.3 & -1.179 & 98.1 \\
\textbf{\texttt{[2]}} & Synthetic Finetuned  & 21.9 & 0.061 & 91.9 & 28.2 & 0.305 & 89.1 & 15.0 & -0.111 & 83.9 & 20.2 & 0.291 & 53.9 \\
\textbf{\texttt{[1]}} & Two-pass Finetuned & 36.5 & 0.301 & \textbf{98.0} & \textbf{38.3} & 0.524 & 96.4 & 23.5 & 0.263 & 91.9 & \textbf{29.6} & 0.455 & \textbf{99.1} \\
\bottomrule
 \end{tabular}
 }
\caption{Averaged formal and informal results on the official test split in the \textit{supervised} setting. Best scores from multilingual and bilingual systems are \textbf{bolded}. Our official submissions to the shared task are numbered \textbf{\texttt{[1-4]}}.}
\label{tab:official_average} 
\end{table*}

\begin{table*}[h]
\centering
 \setlength\tabcolsep{2pt}
\scalebox{0.80}{
\begin{tabular}{llrrrrrrcrrrrrr}
 \toprule
 &  \multirow{3}{*}{\textbf{\textsc{Model}}} &  \multicolumn{6}{c}{\cellcolor{gray!10}\textbf{To Formal}} & &
  \multicolumn{6}{c}{\cellcolor{gray!10}\textbf{To Informal}} \\
 &   &  \multicolumn{3}{c}{\textbf{\en-\ita}}  & \multicolumn{3}{c}{\textbf{\en-\ru}} &&
  \multicolumn{3}{c}{\textbf{\en-\ita}} &
  \multicolumn{3}{c}{\textbf{\en-\ru}} \\ 
&  & \bleu & \comet & \taskscorer & \bleu & \comet & \taskscorer &&  \bleu & \comet & \taskscorer & \bleu & \comet & \taskscorer \\
     \midrule
& Bilingual baselines & \underline{37.0} & \underline{0.557} & \underline{4.5} &  \underline{27.9} & \underline{0.220} & \underline{93.3} &&  \underline{44.2} & \underline{0.618} & \underline{95.5} & \underline{22.0} & \underline{0.169} & \underline{6.7} \\
 \textbf{\texttt{[1]}} & \googlemt (\textsc{ZS}) & 27.6 & 0.306 & 32.8 & 22.7 & 0.123 & \textbf{100.0} && 32.6 & 0.379 & \textbf{97.9} & 17.0 & 0.058 & 1.1\\
\textbf{\texttt{[4]}} &  \mbart (\textsc{ZS}) & \textbf{30.2} & \textbf{0.545} & 38.7 &  \textbf{26.2} &  \textbf{0.275} & \textbf{100.0} && \textbf{35.0} & \textbf{0.597} & 95.9 & \textbf{20.8} & \textbf{0.203} & \textbf{13.8}  \\ 
\textbf{\texttt{[5]}} &  \googlemt  (\textsc{FS}) & 27.1 & 0.302 & \textbf{49.7} & 20.7 & 0.007 & \textbf{100.0}  & & 31.2 & 0.346 & 93.3 & 15.5 & -0.050 & 1.8  \\
\bottomrule
 \end{tabular}
 }
\caption{ Results on  the official test split for the \textit{zero-shot} setting. Our official submissions to the shared task are numbered \textbf{\texttt{[1-5]}}.}
\label{tab:official_zeroshot} 
\vspace{-0.5cm}
\end{table*}

\section{Official Results}

\rw{For tables 8 and 9, it might be better to combine them into one table, since the idea of the shared task is to have one system that can work for both formality levels. I don't think it's very informative to say that, for example, formality agnostic systems works better for EN-JA formal and gold fine-tuned for EN-JA informal; what would be informative is to make it clear which system is better at *controlling* formality according to the desired level.}
\sa{Instead of changing the entire discussion, maybe we could add a paragraph that discusses the controllability in both directions. Elijah added a table including the average of scores from formal/informal directions, which I think solves the purpose. TODO: update discussion}

\paragraph{Submissions}  We submit five variants of multilingual models (numbered \textbf{\texttt{[1-5]}} in Tables~\ref{tab:official_formal}-\ref{tab:official_zeroshot}), and compare them to the bilingual models built on top of the organizers' baselines. We first discuss results on the official test split for the \textit{supervised} setting (Tables~\ref{tab:official_formal}{, }\ref{tab:official_informal}). To better understand the degree of overall control afforded, we also report the average scores of the formal and informal settings in Table ~\ref{tab:official_average} before turning to the \textit{zero-shot} setting in Table~\ref{tab:official_zeroshot}.

\paragraph{Multilingual Approach} The best multilingual models (\texttt{[1]} \& \texttt{[4]}) consistently outperform the {\ttfamily bilingual formality-agnostic} baselines, improving both translation quality (Worst-case gain in Average \bleu: Formal ($+1.67$), Informal: ($+3.7$)) and formality accuracy (Worst-case gain in Average \taskscorer: Formal ($+40.38$), Informal: ($+31.6$)). They approach the quality of formal and informal bilingual systems, but the gap in translation quality and formality accuracy varies across languages. 
While for \de\xspace and \es, there is a large difference in translation quality (approx. 10 \bleu points) between the multilingual models and the bilingual baselines, the multilingual models consistently get higher formality accuracy across language pairs and style directions and also perform comparably with the bilingual models in matching the translation quality for \hi and \ja. We attribute these differences to the amount of training data used across the language pairs (\hi: $0.7$M to \de $20$M). This is an encouraging result, since the bilingual approach uses a much larger language-specific parameter budget and bitext for training than the all purpose multilingual models, which can benefit from transfer learning across languages. 

\paragraph{m\textsc{BART} vs. m\textsc{T5}} The {\ttfamily gold finetuned \mbart} model achieves the best overall translation quality among the multilingual variants as expected given that \mbart is pre-trained on parallel text. Its translation quality is higher than that of \googlemt models according to \bleu and \comet for all languages except 
\hi (informal), which could be attributed to the nature and amount of pre-training data used for \hi. Its formality accuracy is in the $90$'s and within $5$ percentage points to the highest score for all languages except Japanese ($78.2$\%) in the formal direction. In the informal direction, the gap between \mbart and the best system on formality accuracy is larger across the board (Average Acc.: $+19.3$), suggesting that finetuning on gold data cannot completely recover an informal translation despite generally strong performance in formal translations.

\paragraph{Finetuning strategies} Results show that the combination of synthetic and gold data is crucial to help the \googlemt-based model learn to translate and mark formality appropriately.
Finetuning only on the gold data leads to overfitting: the model achieves high formality accuracy scores, but poor translation quality (\bleu $<$ 10). Manual inspection of \googlemt-based system outputs suggests that translations often include tokens in the wrong language (Appendix Table ~\ref{tab:example_output}).
Finetuning on synthetic data improves translation quality substantially compared to gold data only (Average gain in \bleu: Formal ($+15.8$), Informal ($+14.6$)). Two-pass finetuning improves formality control (Average gain in \taskscorer: Formal ($+5.43$), Informal ($+27.85$)), with additional translation quality improvement across the board over synthetic-finetuned model (Average gain in \bleu: Formal ($+10.27$), Informal ($+11.03$); \comet: Formal ($+0.247$), Informal ($+0.252$)). While we primarily focused on the impact of synthetic supervision on \googlemt, we believe a similar investigation using \mbart would yield interesting results and leave this as future work.

\paragraph{Performance across languages} While the higher resource language pairs (\de, \es) achieve better translation quality (in \bleu and \comet) over the relatively lower resource languages (\hi, \ja), the formality accuracy is more comparable across the language pairs for the multilingual models (standard deviation: \googlemt (4), \mbart (10)).  
We can observe that the task accuracy is lowest ($<90$\%) when translating to formal Japanese. By inspection, we observe three broad classes of errors: 1) lexical choice, 2) cross-script matching, 3) ambiguity in politeness levels \cite{feely-etal-2019-controlling}. Lexical choice is invariant in machine translation and is occasionally a valid error in the case of mistranslation, so we focus on the latter two error cases. Japanese has three writing systems and false positives in formality evaluation can occur when surface forms do not match as in the case of \begin{CJK*}{UTF8}{min}面白い\end{CJK*} which can also be written as \begin{CJK*}{UTF8}{min}おもしろい\end{CJK*} (gloss: `interesting'). Finally, there are cases in which the system and reference formality mismatch but can both be interpreted as formal (e.g.,  \begin{CJK*}{UTF8}{min}働きます\end{CJK*} vs. \begin{CJK*}{UTF8}{min}働く\end{CJK*}; gloss: `work' (polite) vs. `work' (formal)).

\rw{In the zero-shot setting, do you use any cross-lingual *formality-specific* data? For example, do you use the formality-labeled data for EN-ES to help the EN-IT model? Or are the multilingual models the only cross-lingual aspect?} \sa{We do not study this in our current work, so I am not sure if it is still helpful to include any clarification or footnote regarding the same.}
\paragraph{Zero-Shot} We observe limited zero-shot transfer of grammatical formality to unseen languages (Table~\ref{tab:official_zeroshot}). For both {\ttfamily \mbart} and {\ttfamily \googlemt} models, the \en-\ita performance is biased towards informal translations, while \en-\ru is biased in the formal direction. In the case of \en-\ita, both \mbart and \googlemt almost always interpret the English second person pronoun as second person \textit{plural} when translating to formal, exploiting the ambiguity of English on the source side. By contrast, when generating informal translations, pronouns are typically preserved as singular. In comparison, with \googlemt-based translations into \ru, we see almost unanimous preference toward the formal, likely due to sampling bias when curating the synthetic training set. We also observe that \mbart prefers to translate in a formal manner irrespective of desired target. In addition, when \mbart fails to account for the target formality, it often generates paraphrases of the formal target. These strong preferences might be symptoms of systematic differences in formality across languages in the training data of these models. 
Finally, the use of silver standard formality labels (``fully supervised'' setting (\textsc{FS})) does not improve over the zero-shot approach, with similar observations of \googlemt-based translations as outlined above. We observe that in the case of \en-\ru, there is a higher incidence of code-switched translations. This may indicate noise introduced in the automatic labeling process and requires further examination in future work.
%

\section{Related Work}

 Most \mt approaches only indirectly capture the style properties of the target text. While efforts have been made to generate better outputs in their pragmatic context via controlling formality \cite{sennrich-etal-2016-controlling,feely-etal-2019-controlling, niu2020controlling, schioppa-etal-2021-controlling}, complexity \cite{marchisio-etal-2019-controlling, agrawal-carpuat-2019-controlling}, gender \cite{rabinovich-etal-2017-personalized}, these studies only focus a single language pair.
 Due to the paucity of style annotated corpora, zero-shot style transfer within and across languages has received a lot of attention.  However, adapting pre-trained large-scale language models during inference using only a few examples \cite{garcia-etal-2021-towards, riley-etal-2021-textsettr, krishna2022fewshot} limits their transfer ability and the diversity of their outputs. While prior works use pre-trained language models like \bert, \textsc{GPT} to intialize $\theta_{LM}$ for improving translation quality \cite{guo2020incorporating, zhu2019incorporating}, in this work, we focus on adapting sequence-to-sequence multilingual models for controlled generation of a desired formality and study style transfer in multilingual supervised and zero-shot settings.




%

\section{Conclusion}

We present the University of Maryland's submission which examines the performance of a single multilingual model allowing control of both target language and formality. Results show that while multilingual \fsmt models lag behind large, bilingual, formality-specific models in terms of MT quality, they show stronger formality control performance across all the language pairs.
Furthermore, while synthetic unpaired triplets help {\ttfamily \googlemt} with \fsmt performance and the two-stage finetuning process improves \mt quality and contrastive task performance, {\ttfamily \mbart} still outperforms this class of models, likely due to its large amount of pre-training supervision.

In future work, we suggest a deeper investigation of potentially confounding roles in the study of \fsmt, such as the impact of formal register as compared to grammatical formality in training data. We also suggest a thorough analysis of \textit{what} is transferred in the zero-shot setting. Finally, we recommend an audit of underlying pre-training and finetuning data sources for pre-trained multilingual models, which we believe hinder zero-shot formality transfer for \en-\ita and \en-\ru in which a single formality is strongly preferred.

\bibliography{anthology,custom}
\bibliographystyle{acl_natbib}

\appendix
\newpage
\onecolumn
\label{sec:rules}
\section{Rules for Synthetic Data Curation}

\sa{Added for camera ready: } 

\begin{table*}[h]
\scalebox{0.90}{
\centering
\begin{tabular}{lll}
\toprule
\textsc{Lang}  & Formal & Informal \\
\midrule
en-de & (P=2 $\in$ M and Num=Plural $\in$ M) or PP=Sie & P=2 $\in$ M and Num=Plural $\notin$ M\\
en-es & P=2 $\in$ M and Form=Polite $\in$ M  &  P=2 $\in$ M and Num=Singular $\in$ M and Form=Polite $\notin$ M   \\
en-it & PP=voi or PP=lei & PP=tu \\
en-ru & PP=\foreignlanguage{russian}{Вы} & PP=\foreignlanguage{russian}{ты} \\
\bottomrule
    \end{tabular}}
    \caption{Rules for extracting formal and informal sentences for each language pair from existing bitext. P: Person; PP: Personal pronoun; N: Number; $x \in M$ indicates that some token within the sentence has morphological features matching $x$ as produced by spaCy. }
\label{tab:rules} 
\end{table*}

\section{Glosses}
\label{sec:glosses}
\subsection{Necessarily formal}

Appropriate pronouns with accompanying conjugation imply the sentence is grammatically formal.

\exg. ¿Cuándo nació usted? \hfill (Spanish)\\
        When born {you (form.)}?\\
    \glt `When were you (form.) born?'

\exg. Woher  kommen Sie? \hfill (German)\\
        {Where from} come {you (form.)?}\\
    \glt `Where are you (form.) from?'

\subsection{Necessarily informal}

Appropriate pronouns with accompanying conjugation imply the sentence is grammatically informal. Note that Spanish is pro-drop, which relaxes the requirement on personal pronouns.

\exg. ¿Cuándo naciste (tú)? \hfill (Spanish)\\
        When born {you (inf.)}?\\
    \glt `When were you (inf.) born?'

\exg. Woher kommst du? \hfill (German)\\
        {Where from} come {you (inf.)?} \\
    \glt `Where are you (inf.) from?'

\subsection{Ambiguously formal}

Because Spanish is pro-drop, personal pronouns can be omitted depending on context. Since formal conjugations are shared with neutral third person subjects, this leaves ambiguity when the pronoun is dropped. For sake of gloss, we use $\emptyset$ to indicate a dropped pronoun.

\exg. ¿Cuándo nació $\emptyset$? \\
        When born {\{you (form.), he, she, it\}?}\\
    \glt `When \{were you (form.), was \{he, she, it\}\} born?'


\section{Official Evaluation} \label{sec:eval}

\begin{figure*}[t]
\begin{subfigure}{0.45\textwidth}
  \centering
  \includegraphics[width=\linewidth]{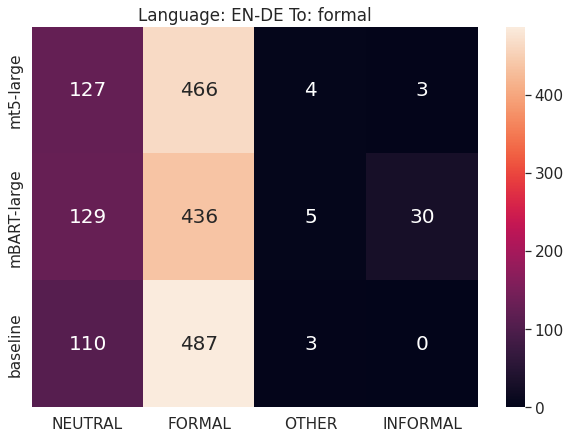}
  \caption{} \label{a}
\end{subfigure}%
\begin{subfigure}{0.45\textwidth}
  \includegraphics[width=\linewidth]{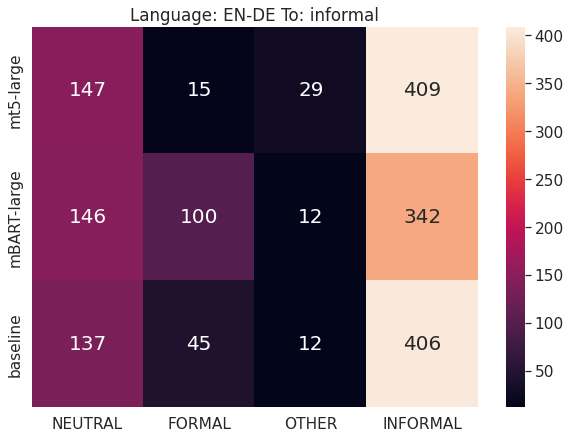}
   \caption{} \label{b}
\end{subfigure}
\begin{subfigure}{0.45\textwidth}
  \centering
  \includegraphics[width=\linewidth]{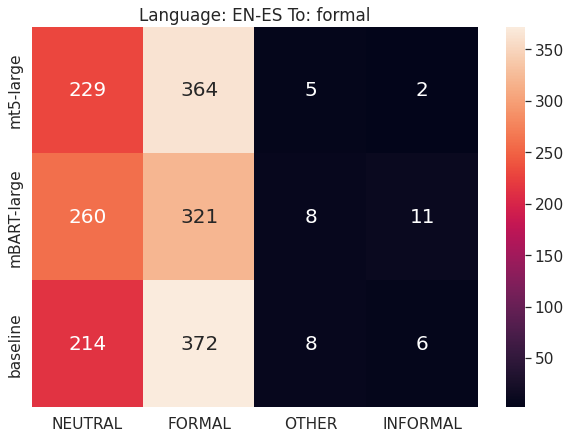}
  \caption{} \label{a}
\end{subfigure}%
\begin{subfigure}{0.45\textwidth}
  \includegraphics[width=\linewidth]{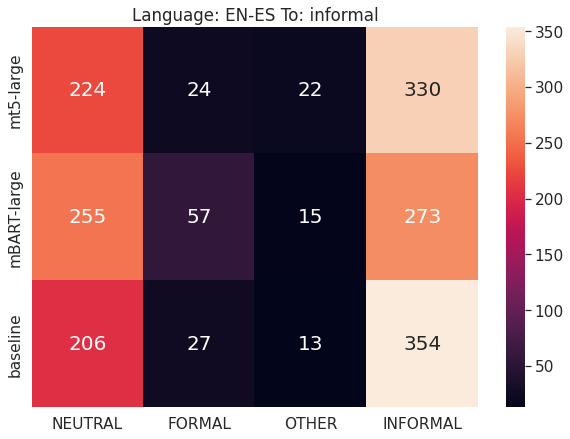}
   \caption{} \label{b}
\end{subfigure}
\begin{subfigure}{0.45\textwidth}
  \centering
  \includegraphics[width=\linewidth]{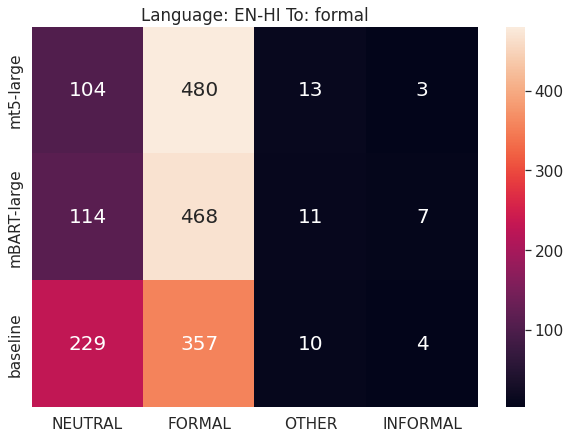}
  \caption{} \label{a}
\end{subfigure}%
\begin{subfigure}{0.45\textwidth}
  \includegraphics[width=\linewidth]{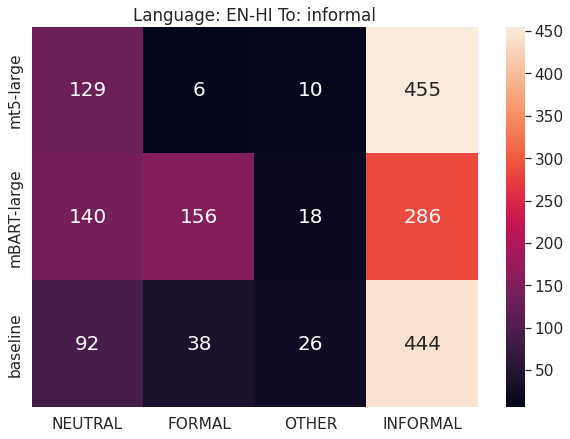}
   \caption{} \label{b}
\end{subfigure}
\begin{subfigure}{0.45\textwidth}
  \centering
  \includegraphics[width=\linewidth]{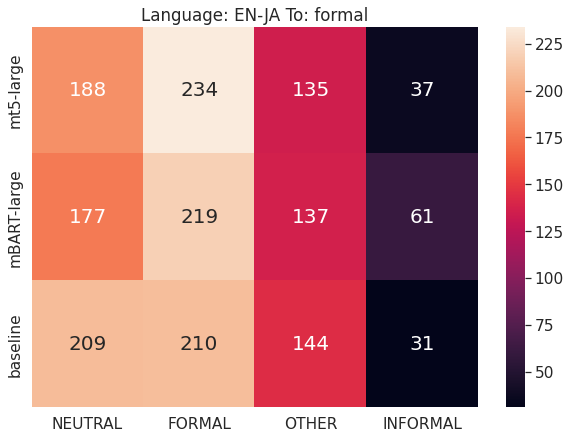}
  \caption{} \label{a}
\end{subfigure}%
\begin{subfigure}{0.45\textwidth}
  \includegraphics[width=\linewidth]{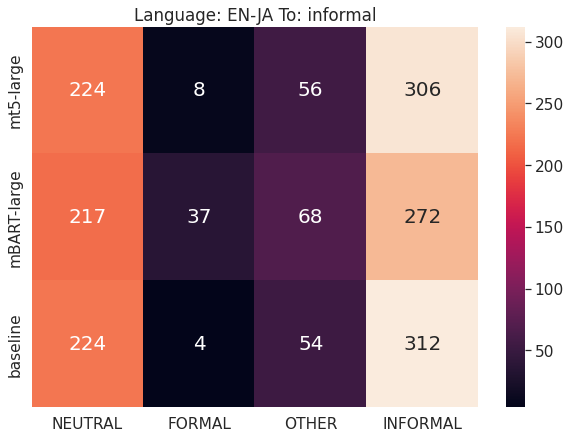}
   \caption{} \label{b}
\end{subfigure}
\caption{Class Distribution for the baseline, \mbart and \googlemt systems for all the supervised language pairs.}
\label{fig:eval}
\end{figure*}

We report the number of examples labeled as \textsc{FORMAL}, \textsc{INFORMAL}, \textsc{NEUTRAL}, \textsc{OTHER} by the formality scorer for the best multilingual models ( \textbf{\texttt{[1, 4]}}) and the baseline systems for each language pair and formality direction. As described in \ref{metrics}, the accuracy is computed based on \textit{realized} matches, which excludes examples labelled as \textsc{NEUTRAL} and \textsc{OTHER}. Figure~\ref{fig:eval} shows that the number of these excluded \textsc{NEUTRAL} samples can range from $15$\% to $43$\%.

\newpage

\section{Example Outputs}

\begin{table}[ht]
    \centering
        \renewcommand\tabularxcolumn[1]{m{#1}}
        \renewcommand\arraystretch{1.3}
    \begin{tabularx}{\columnwidth}{*{1}{>{\arraybackslash}X}}
  \textbf{Source:} Wow, that's awesome! Who is your favorite Baseball team? I like my Az team lol \\
  \textbf{German Formal Hypothesis:} Wow, das ist toll! Wer ist Ihr Lieblings- Baseballteam? Ich mag meine Az-Team lol. \\
  \textbf{German Formal Reference:} Wow, das ist fantastisch! Welches ist Ihr Lieblingsbaseballteam? Ich stehe auf mein AZ-Team lol. \\
  \textbf{German Informal Hypothesis:} Wow, das ist toll! Wer ist dein Lieblings\begin{CJK*}{UTF8}{gbsn}野球\end{CJK*}team? Ich mag meine Az Team lol. \\
  \textbf{German Informal Reference:} Wow, das ist fantastisch! Welches ist dein Lieblingsbaseballteam? Ich stehe auf mein AZ-Team lol. \\  
    \end{tabularx}  
    \caption{Contrastive outputs from English-German. Note that there is not only variety in lexical choice between references and hypotheses, but also between hypotheses of varying formality (i.e., \begin{CJK*}{UTF8}{min}野球\end{CJK*} is ``baseball'' in Japanese)} \label{tab:example_output}
    \vspace{-0.5cm}
\end{table}

\section{Accuracy of Formality Classifiers}

\sa{Added for camera ready: } 
We report the accuracy of the learned classifiers on the \internaltrain dataset in Table~\ref{tab:accuracy}.  
\begin{table}[h]
\centering
\begin{tabular}{lrr}
\toprule
\multirow{2}{*}{\textbf{\textsc{Language}}}  & \multicolumn{2}{c}{\textbf{Accuracy}} \\
& Formal & Informal  \\ \midrule
en-de &  98\% & 99\%    \\
en-es &  99\% & 92\%   \\ 
en-ja &  98\% & 98\%      \\ 
en-hi &   96\% & 95\%                        \\                          \bottomrule
\end{tabular}
\caption{Accuracy of trained formality classifiers on the \internaldev dataset.}
\label{tab:accuracy} 
\end{table}

\end{document}